# JRC-NAMES: A Freely Available, Highly Multilingual Named Entity Resource

**Ralf Steinberger, Bruno Pouliquen, Mijail Kabadjov,
Jenya Belyaeva & Erik van der Goot**

European Commission – Joint Research Centre
Via Enrico Fermi 2749, 21027 Ispra (VA), Italy
`{Firstname.Lastname}`@jrc.ec.europa.eu

## Abstract

This paper describes a new, freely available, highly multilingual named entity resource for person and organisation names that has been compiled over seven years of large-scale multilingual news analysis combined with Wikipedia mining, resulting in 205,000 person and organisation names plus about the same number of spelling variants written in over 20 different scripts and in many more languages. This resource, produced as part of the *Europe Media Monitor* activity (EMM, http://emm.newsbrief.eu/overview.html), can be used for a number of purposes. These include improving name search in databases or on the internet, seeding machine learning systems to learn named entity recognition rules, improve machine translation results, and more. We describe here how this resource was created; we give statistics on its current size; we address the issue of morphological inflection; and we give details regarding its functionality. Updates to this resource will be made available daily.

## 1 Introduction

The release consists of named entity lists and Java-implemented software. The software performs two main functionalities: (1) It recognises known names in text of any language and returns the name as it was found, its position and length, the standard variant of the name and the unique numerical name identifier. (2) It allows exporting all known name variants so that users can exploit the resource in further ways.

*JRC-Names* is the result of a multi-year large-scale effort to recognise new person and organisation names in up to 100,000 news articles per day in up to 20 different languages[1], to automatically recognise which newly found names are variants of each other, to enhance the name list with additional information extracted from Wikipedia, and to manually improve the database entries for the most frequently found names. While the Named Entity Recognition (NER) module itself is not part of the release, updates to *JRC-Names* will be made available daily so that users will always have access to the latest named entity (NE) list.[2]

In the following sections, we describe possible uses of this resource (2), list other available NE resources (3), explain very briefly how the resource was built (4), give some statistics on the resource and provide technical details on the released software (5).

## 2 Uses of this named entity resource

The tool serves many purposes and addresses various problems, including the following:

(a) Proper names are a problem when searching databases, the internet and other repositories, because variants of searched names are often not found (Stern & Sagot 2010). This results in non-optimal use and exploitation of repositories for documents, images and audio-visual content. *JRC-Names* allows standardising the names and improving retrieval;

(b) Names are a known problem for machine translation as they should not be translated like other words (Babych & Hartley 2003); names can be extracted before the translation process and the foreign language variant can be re-inserted in the target language to solve this problem;

---

[1] EMM-NewsExplorer (http://emm.newsexplorer.eu/) currently extracts new names from news articles in Arabic, Bulgarian, Danish, Dutch, English, Estonian, Farsi, French, German, Italian, Norwegian, Polish, Portuguese, Romanian, Russian, Slovene, Spanish, Swahili, Swedish and Turkish.

[2] Accessible via http://langtech.jrc.ec.europa.eu/.

(c) Lists of names in two different scripts are often used to learn transliteration rules (e.g. Pouliquen 2009).
(d) Names can be recognised and marked up in text to use as seeds when training a machine learning NER system (e.g. Buchholz & van den Bosch 2000);
(e) Social networks are less biased by national viewpoints if produced using multi-national sources and entity lists;
(f) Recognition of names is useful as input to the tasks of opinion mining, co-reference resolution, summarisation, topic detection and tracking, cross-lingual linking of related documents across languages, etc.

*JRC-Names* is a resource that can be useful in all these scenarios. Potential beneficiaries of the tool are IT developers and researchers in the field of text mining and machine translation; news agencies, photo agencies and other media organisations; business intelligence, and possibly more.

## 3 Related work

In this section, we summarise previous efforts to compile multilingual name lists. Work on developing NER systems is abundant and shall not be discussed here. For an overview of the state-of-the-art in NER, see Nadeau & Sekine (2009).

Wentland et al. (2008) built a multilingual named entity dictionary by mining Wikipedia and exploiting various link types. They first built an English named entity repository of about 1.5 million names, by selecting all article headers and by assuming that these headers are named entities if at least 75% of these strings are more frequently found in uppercase than in lowercase (except at the beginning of a sentence). They then exploit the multilingual links, as well as the redirect and disambiguation pages to identify target language equivalences in altogether fifteen languages. This method produced 250,000 named entities for the most successful language German, and about 3,000 for the lesser-resourced language Swahili.

Toral et al. (2008) built the resource *Named Entity WordNet* by searching for NEs in WordNet and by combining information found in WordNet and in Wikipedia. The resource consists of 310,000 entities, including 278,000 persons. Name variants found in Wikipedia are included.

Stern & Sagot (2010) exploit Wikipedia and GeoNames to produce a French language-focused NE database which includes 263,000 person names and 883,000 variants, extracted from French Wikipedia entries by parsing the first sentence and the redirection pages.

Prolexbase (Maurel, 2008) is a mostly manually produced resource containing about 75,000 names for different entity types, built up over many years.

With the exception of Prolexbase, all of these person name resources are the result of exploiting Wikipedia. Wikipedia is strong at providing cross-lingual and cross-script variants, but it contains only few spelling variants within the same language and it does not contain information on morphological variants. In contrast, our resource is mostly built up by recognising name variants in real-life multilingual text, and it additionally contains Wikipedia variants, resulting in up to 400 spelling variations for a single name. Future releases of *JRC-Names* will also recognise morphological inflections of entity names.

## 4 How the NE resource was created

This section summarises the role of NER in EMM (Section 4.1), explains how the NE information in *JRC-Names* is extracted (4.2), how the tool automatically detects which name strings are variant spellings for the same entity (4.3), and how the NE database is enhanced through human moderation (4.4) and with Wikipedia (4.5), and how morphological variants are recognised (4.6). Due to space limitations, the sub-sections on NER (4.2) and on name variant merging (4.3) are very brief, but this work has been described in much detail in Steinberger & Pouliquen (2009).

### 4.1 Role of NER in EMM

The freely accessible *Europe Media Monitor* (EMM) family of applications gather a current average of 100,000 news articles per day in up to 50 languages from the internet, classify them into hundreds of categories, cluster related news, link news clusters over time and across languages, and – for twenty languages – perform entity recognition, classification and disambiguation for the entity types person, organisation and location. EMM also gathers information about entities from all news articles and displays it on over one million entity pages. For an overview of EMM, see Steinberger et al. (2009).

### 4.2 Multilingual NER from the news

NER in EMM is performed using manually constructed language-independent rules that make use of language-specific lists of titles and other

معمر القذافي; Mouammar Kadhafi; Muammar al-Gaddafi; Moammar Gadhafi; Muammar Gheddafi; Муамар Кадафи; Muammar Kadhafi; Muammar Kaddafi; Muammer Kaddafi; Muamar Gadafi; معمر قذافي; Moamerja Gadafija; Muammar Kadafi; Muammar el Gaddafi; Муамар Каддафи; Muamar el Gadafi; Moammar Gaddafi; Moamar Gaddafi; Moamer Kadhafi; Muammar Gadafi; Moamer Gadafi; Mouammar Khadafi; Moamar Kadhafi; Muammar Gadaffi; Muammar Khadaffi; Muammar Khaddafi; Muammar Qaddafi; Muhammar Gheddafi; Muammar al Gaddafi; Moammar Gaddafi; Muammar Kadafi; Муаммар Каддафи; Moamer Gathafi; Muammar Khadafi; Mouammar al-Kadhafi; Muamar al Gadafi; Muammar el-Qaddafi; Muammar Gadafy; Muammar Kadaffi; Muammer Gadhafi; Moamer Gaddafi; Muammar al-Ghadhafi; Muamar Gaddafi; Muammar Ghadafi; Muamar Khadafi; Muammar Ghadhafi; Muammar al-Gadafi; Muammar al-Qadhafi; Mouammar El Kadhafi; Muammar Qadhafi; Muammer Gadaffi; Moammar Gheddafi; Mouamar Kadhafi; Mouamar Khadafi; Moamer Kadafi; Moammar al-Qadhafi; Moamer Qadhafi; Moamar Kadhafi; Moammar Khadafi; Moamar Gadafi; Moamar Qaddafi; Muammer Gaddafi; Muammar el-Gaddafi; Moeammar Kadhafi; Mummar Gaddafi; Muammar al-Qathafi; Muammar al-Kadhafi; Muammar Al-Kaddafi; Muammar Al-Qadhafi; Moammar Khadafi; Muammar al-Qaddafi; Mouammar Al Kadhafi; Moammar Ghadafi; Muammar Al Gaddafi; Muammar Kaddafi; Moammar al-Kadhafi; Muammar El-Kadhafi; Moammar Khaddafi; Moammar Qadhafi; Muammar al-Gathafi; Muammar Ghadaffi; Muhammar Gaddafi; Muammar Gaddaffi; Muammar el Gadafi; Muammar Abu Minyar al-Gaddafi; Muammar al-Kadafi; Muhamar Kadafi; Mouamar Kaddafi; Moammer Gaddafi; Muammar Al-Gaddafi; Muammar al-Khadafi; Mouammar El Khaddafi; Muammar Gadhaffi; Моамар Кадафи; Muamar Al Gadafi; Mouammar words and phrases that are typically found next to names, such as titles (*president*), professions or occupations (*tennis player, playboy*), references to countries, regions, ethnic or religious groups (*French, Bavarian, Berber, Muslim*), age expressions (*57-year-old*), verbal phrases (*deceased*), modifiers (*former*) and more. These pattern words, which we refer to as *trigger words*, can also occur in combination (*57-year-old former British Prime Minister*) and patterns can be nested to capture more complex titles (e.g. *current Chair of RANLP* Ruslan Mitkov). In order to be able to cover many different languages, no other dictionaries and no parsers or part-of-speech taggers are used. To avoid detecting strings such as *Monday Angela Merkel* as a name, non-name uppercase words (including *Monday*) from a *name stop word* list are excluded from the recognition. The trigger word files contain between a few hundred and a few thousand words and regular expressions per language (to deal with inflection and other variations). Trigger word lists are produced in a combination of a manual collection from various online sources, machine learning and bootstrapping. The trigger words found historically next to each name are stored in order to build up a frequency-ranked repository of common titles (and more) for each entity.

**Figure 1.** Name variant spellings for Libyan leader Muammar Gaddafi, as found in multilingual media reports.

The method is relatively simple and may at first seem labour-intensive, compared to machine learning methods that learn recognition rules on the basis of examples. However, its main advantages are that it is light-weight (it does not require linguistic tools such as morphological analysers or part-of-speech taggers), it is modular (a new language can simply be plugged in by providing the language-specific trigger words, rules and trigger word lists can be manually verified and corrected so that the method allows high levels of control. Further details can be found in Steinberger & Pouliquen (2009).

Organisation name recognition is relatively weakly developed in EMM's media monitoring applications. Organisation names are recognised if one of the words of the name candidate is a typical organisation name part from a given list (*organisation, club, international, bank*, etc.). Additionally, a Bayesian classifier trained on lists of known person and organisation names decides on the type of a new entity. Due to our coarse entity type categorisation, other entity types are frequently included into the type *organisation*, such as *Belfast Agreement*, *Nobel Prize*, *Red Mosque* or *World War I*. The entity type *Organisation* should thus be interpreted as *Other Entities*.

### 4.3 Name variant matching

The NER tool identifies about 1,000 new names per day and the name database currently contains about 1.15 million different entities plus about 200,000 additional spelling variants (see **Figure 1** for an example of naturally occurring spelling variants). To identify which of the names newly found every day are new entities and which ones are merely variant spellings of entities already contained in the database, we apply a language-independent name similarity measure to decide which name variants should be automatically merged. This algorithm carries out the following steps, which are the same for all languages and scripts: (1) If the name is not written using the Roman script: Transliteration into the Roman script (using standard n-to-n character transliteration rules); all names are lower-cased; (2) name normalisation; (3) vowel removal to create a consonant signature; (4) for all names with the same consonant signature, calculate the overall similarity between each pair of names, based on the edit distance of two representations of both names: between the output of steps (1) and (2). If the overall similarity of two names is above the empirically defined threshold of 0.94, the two names are automati-cally merged. If the similarity lies below that value, they are kept as separate entities. This threshold was set to reach almost 100% precision and to avoid erroneously merging variants of different entities. Additional variants can be assigned to known names in a manual verification process (see 4.4).

The normalisation rules (see **Figure 2**) are hand-crafted, based on the observation of regular name spelling variations. The method for normalisation and variant mapping is the same for all languages and all rules apply to all languages.

### 4.4 Daily manual verification and improvement

The process described in the previous sections does not as such need manual intervention, but human control does help improve the quality of the database regarding a number of issues: (1) correct recognition mistakes such as *Genius Report* or *Opfer von Diskriminierung* (English: *Victim of Discrimination*) – such names will be kept to avoid their renewed recognition in the future; (2) tune the NER process (e.g. by adding newly found name stop words, such as *Report*); (3) merge name variants whose similarity lies below the merger threshold; (4) change the main name of an entity; (5) correct the entity type (person P, organisation O, toponym T); (6) launch an automatic Wikipedia mining process (see 4.5). Manual intervention is only carried out for the most frequently mentioned names, or for regular mistakes that affect large numbers of entities (e.g. weekdays being recognised as part of the name; or morphological inflections erroneously being recognised as regular name variants). Due to high user visibility, the manual process additionally focuses on entities involved in large events such as the Olympics, Oscar and Nobel Prize nominations, and similar. An average of one hour of human effort per day is currently dedicated to these tasks.

### 4.5 Wikipedia lookup to add name variants in more languages

An automatic routine allows the human moderator to retrieve from Wikipedia additional name variants, as well as a photograph for any given entity, if available. The tool checks – for all known name variants of an entity – whether a Wikipedia entry exists and, if successful, mines the cross-lingual links for additional multilingual name variants. It is due to this process that the database contains name variants in languages for which EMM's NER tool has not yet been developed (e.g. Chinese, Japanese and Hebrew). The Wikipedia mining process is not launched in batch mode to allow verifying the correctness of the photograph.

### 4.6 Morphological inflections of names

In many languages, proper names and other words are morphologically inflected. Adding inflected names to the database would be inefficient and untidy. At the same time, simple lookup procedures would miss inflected names when only searching for the base form. In order to capture at least a large part of the inflected names, inflections for names having been found in at least five different news clusters are pre-generated, separately for each language, for all known name variants. The rules for the morphological expansion are hand-crafted, following the major morphological patterns of a language (see **Figure 3**). They do not cover all exceptions, and they may over-generate, i.e. produce forms that do not actually exist in that language. However, they are very efficient and they allow to recognise a majority of name inflections in text and to return the base form for that name.

In addition to morphological variants, this method also produces other regular variants: For instance, non-hyphenated variants of

Latin normalisation:
- accented character → non-accented equivalent — Malik al-Saïdoullaïev → Malik al-Saidoullaiev
- double consonant → single consonant — Malik al-Saidoulaiev
- ou → u — Malik al-Saidulaiev
- " al-" → — Malik Saidulaiev
- wl (beginning of name) → vl — ... mlk sdlv
- ow (end of name) → ov
- ck → k
- ph → f
- ž → j
- š → sh
- x → ks
- Remove vowels

| Name | Normalised form |
|---|---|
| Mohammed Siad Barre, Mohamed Siad Barré, Мохаммед Сиад Барре, محمد سياد بري | **m hm d sd br** (mohamed siad bare) |
| Mahmoud Ahmadinejad, Mahmūd Ahmadīnežād | **m hm d hm dnjd** (mahmud ahmadinejad) |

**Figure 2.** Selection of name normalisation rules and their result. The hand-crafted rules are based on empirical observations about regular spelling variations. They are purely pragmatically motivated and not intended to represent any linguistic reality.

```
Tony(a|o|u|om|em|m|ju|jem|ja)?\s+Blair(a|o|u|
om|em|m|ju|jem|ja)
```

**Figure 3.** Regular expression for the automatic creation of Slovene inflection forms for the name of the former British Prime Minister *Tony Blair*.

hyphenated names are being pre-generated, and Arabic names without the name particles *al* or *el* when the full name contains them, etc. (e.g. *Mohammed al-Mahdi --> Mohammed Mahdi*).

## 5 Statistics and technical details

We will first give details about EMM's entire name database (Section 5.1) and then about the subset of data that is part of the *JRC-Names* distribution (5.2). The remaining sub-sections explain how to read the NE resource file (5.3), give details about the accompanying software (5.4) and discuss plans for future extensions of JRC-Names (5.5).

### 5.1 Some statistics on the name database

EMM's NE database currently contains 1.18 million person and 6,700 organisation names (status July 2011). Additionally, it contains about 200,000 person and 25,000 organisation name variants. The database grows by almost 1,000 name forms (names or variants) per day. The names in the database are written in 27 different scripts (See **Table 1** for the top of the frequency list, ranked by names including their variants). Latin includes all European Union languages except Greek and Bulgarian; the Arabic script also covers Farsi.

The question regarding the distribution of the names across different languages is not easy to answer as the news tends to mention names from around the world. The fact that a certain name is more often mentioned by the press in one country (or in combination with a certain country name) can be misleading. For instance, entity number 10101 (*European Union*) was the most frequently mentioned entity in German language news in 2010 (before *Angela Merkel*) and it was the second most frequently mentioned name in English language news (after *Barack Obama*). However, for look-up purposes in most European languages, it does not matter whether *Silvio Berlusconi* is an Italian, German or Romanian name, as long as it gets recognised in texts of that language.

### 5.2 Statistics on *JRC-Names*

*JRC-Names* does not contain the entire contents of our database. Instead, it contains the subset of names that satisfy at least one of the following conditions: (a) they have been found in at least five different news clusters; (b) they have been manually validated; (c) they have been retrieved from Wikipedia. The first condition helps to drastically cut down on wrongly identified names. Names thus need to be found repeatedly and in different contexts before they get accepted in the list of *known names*. Secondly, names that have been mentioned only once or twice in the course of many years (they are the majority), will be less useful for most users.

The released data contains about 205,000 distinct names and 204,000 additional variants (status July 2011). The dataset grows by about 230 new entities and an additional 430 new name variants per week. The data set contains relatively few organisations (3.2%). Out of this current total of 205,000 unique names, almost two thirds (63.76%) do not have name variants; 22.52% and 5.31% have two and three variants, respectively. There are 3760 names with ten or more variants, 242 with 50 or more, and 37 with more than 100 variants. The names with the most name variants are *Muammar Gaddafi* (413 variants, see **Figure 1**), *Mikhail Saakashvili* (256 name variants) and *Mahmoud Ahmadinejad* (246 variants).

Only an extremely small subset of these names and their variants has been manually verified (although manual moderation does focus on the most highly visible and the most frequently men-tioned names). For this reason, the name list will contain a number of errors. We have identified the following types: (a) non-entities (e.g. *Red Piano* or *French Doctor*); (b) names with Wikipedia scope notes (e.g. *Vinci (construction)*); (c) names with a wrong name extent (e.g. *Even Obama*); (d) inflected names (e.g. *Tonyjem Blairom*); (e) wrong entity type (e.g. *Merlin Biosciences* as a Person; see the discussion in Section 4.2) and (f) non-unique organisation names (e.g. *Health Ministry*). However, we do not consider spelling mistakes found in media reports (e.g. *Condaleeza Rice* instead of the correct spelling *Condoleezza Rice*) as being errors in JRC-Names as they do occur in real-life texts, and knowing them helps identify intended references to entities. (g) It is unavoidable that several unique identifiers occasionally exist for the same entity (e.g. *Sergei Izvolskij* and *Sergey Izvolskiy*). (h) It is possible that different entities have been merged into one entity. (i) It is furthermore very likely that different persons sharing the same first and last name have the same identifier because no disambiguation mechanism is in place.

| ISO15924 | TEXT | Number Variants | Count Entities |
|---|---|---|---|
| Latn | Latin | 1588622 | 1263969 |
| Cyrl | Cyrillic | 104107 | 88097 |
| Arab | Arabic | 17691 | 14513 |
| Jpan | Japanese (Han+Hiragana+Katakana) | 6995 | 6785 |
| Hans | Han (Simplified variant) | 4751 | 4512 |
| Hebr | Hebrew | 3811 | 3664 |
| Kore | Korean (Hangul+Han) | 2432 | 2354 |
| Deva | Devanagari (Nagari) | 1527 | 1043 |
| Grek | Greek | 1476 | 1410 |
| Thai | Thai | 1203 | 1140 |
| Geor | Georgian (Mkhedruli) | 1072 | 1021 |
| Beng | Bengali | 674 | 645 |
| Taml | Tamil | 639 | 618 |
| Mlym | Malayalam | 278 | 272 |
| Armn | Armenian | 195 | 188 |
| Knda | Kannada | 145 | 139 |
| Telu | Telugu | 128 | 126 |
| Ethi | Ethiopic (Ge︎ez) | 112 | 108 |

**Table 1.** Number of NEs and their variant spellings written in 18 out of the 27 different scripts contained in the NE database.

### 5.3 Reading the named entity resource file

The tool consists of Java-implemented software and a named entity resource file. Updates of the resource file will be made available for download daily so that users will always have the newest NE data. The resource file is a UTF8-encoded Java zip file. There is no need to open this file if the provided software is used, but we describe the file structure here in case users do want to access the file: Each line consists of four tab-separated columns containing: name ID; type; language; and name variant (see **Table 2**). *Name ID* is a unique numerical identifier for the entity. In this release, *Type* can only be Person (P) or Organisation (O). The column *Language* contains the ISO 639-2 two-digit code for the language if the name variant should only be looked up in that language. If a name can be looked up in all languages, which is the default, the value is *u* (undefined). The strings in the column *name variant* are the known spellings of the name, one per line. Multi-word strings are separated by the '+' sign (e.g. *United+Nations*). For all lines with the same name ID, the first line shows the main name, i.e. the variant that we chose to use for display purposes inside EMM. We usually choose it because it either is the name variant most frequently found in the news, or because it is the variant found on Wikipedia, or because it is a frequent Latin script version of a name originally written in another script.

While many name variants will only occur in some languages and not in others, it does not normally do any harm to search for the foreign language variants in a text. However, in some cases, a name variant may have a different meaning in other languages. In such cases, it is useful to restrict the lookup information to a subset of languages, or even to a single language, in order to avoid false positives. To give an example: the short name of the German insurance company *Allianz* is homographic with a common German noun (English *alliance*), so the simple word *Allianz* should not be recognised as the insurance company in German language texts. The multi-word name variants *Versicherer Allianz*, *Allianz SE*, *Allianz-Konzern* and others will be recognised. Another example is the acronym 'FN', which stands for the political party *Front National* in French language text, while it stands for *Förenta nationerna* (*United Nations*) in Swedish. By restricting the lookup to specific languages, we can thus avoid mistakes. See **Table 2** for some sample entries.

```
3202    O    u     United+Nations
3202    O    u     Nations+Unies
3202    O    fr    ONU
3202    O    u     ಸಂಯುಕ್ತ+ರಾಷ್ಟ್ರ+ಸಂಸ್ಥೆ
3202    O    u     Ujedinjeni narodi
3202    O    sv    FN
13752   O    u     Front National
13752   O    fr    FN
13752   O    u     国民戦線
13752   O    u     Фронт Насионал
```

**Table 2.** Selected lines from the name resource file, showing the unique numerical identifier, the entity type, the language scope and the name variant.

### 5.4 Programming details / Usage of the tool

The NE resource file is accompanied by Java code. The code consists of a library that implements the actual text matching, and of a number of source files to demonstrate how to match entities in a text and how to extract the entity information from the NE resource file. This information can then also be used to produce the full list of name variants.

The matching software, after reading and analysing the NE resource file, searches for any of the known entities in multilingual text. For every entity found, the software will return the following values: (a) the numerical name identifier; (b) the main name for that entity; (c) the name string found (this can be any variant from the NE re-source file); (d) the offset in the text; (e) the length of the name string found. The lookup process is case-sensitive: For languages distinguishing case, the uppercase letters in the NE resource file will only match if they are also spelt with uppercase in the text, while lowercase letters will match both upper and lower case.

The tool in principle searches for any of the name variants in texts of any language, with the exception of those cases where names are marked as being language-specific or their recognition is blocked for a specific language, as described in Section 5.3.

The lookup process is fast because the software uses finite state technology. It does not require large amounts of memory. It can be run effortlessly on a modern desktop computer. This tool has been in use for several years. It is robust and its output can be seen on EMM's web pages (see http://emm.newsbrief.eu/overview.html).

### 5.5 Further planned developments

This first release of *JRC-Names* does not recognise morphologically inflected variants of entity names. However, it is planned that future versions will include morphological variant recognition in one of two ways. Either morphological variants will be pre-generated (similar to the process described in Section 4.6) and added to the named entity resource file; or inflection variants will be dealt with as part of the lookup process performed by the software, for instance through the application of regular expressions. The recognition of other name spelling variants will also be included consistently, such as: hyphenated versus non-hyphenated name variants (e.g. *Yves Saint-Laurent* vs. *Yves Saint Laurent*); names with or without name infixes such as 'al' (*Khan al Khalil* vs. *Khan Khalil*); names with and without spaces in languages where space separation is optional (e.g. 巴拉克·歐巴馬 and 巴拉克歐巴馬), and more.

We also plan to make available frequency counts for names and their variants and, if possible, counts of the frequency per language. Furthermore, we may be able to publish the most frequent trigger words (titles and more, see Section 4.2) found next to each entity.

### Acknowledgments

The *Europe Media Monitor* EMM is a multiannual group effort involving many tasks, of which some are much less visible to the outside world. We would thus like to thank all past and present OPTIMA team members for their help and dedication. We would also like to thank our Unit Head Delilah Al Khudhairy for her support.